\documentclass[11pt,journal]{IEEEtran}
\usepackage{blindtext}
\usepackage{graphicx}
\usepackage{cite}
\usepackage[justification=centering]{caption}
\usepackage{lettrine}
\usepackage{etoolbox}
\usepackage{amssymb}
\usepackage{mathtools}
\usepackage{lipsum}

\ifCLASSINFOpdf
\else
\fi

\begin{document}
\title{Object Detection Using Image Processing}

\author{Far\'es~Jalled,~\IEEEmembership{Moscow Institute of Physics \& Technology, Department of Radio Engineering \& Cybernetics~} 

Ilia~Voronkov,~\IEEEmembership{Moscow Institute of Physics \& Technology,  Department of Radio Engineering \& Cybernetics~}

\patchcmd 
\thanks{}
\thanks{\IEEEmembership{}}}

\markboth{}
{Shell \MakeLowercase{\textit{et al.}}: Bare Demo of IEEEtran.cls for Journals}

\maketitle

\begin{abstract}
An Unmanned Ariel vehicle (UAV) has greater importance in the army for border security. The main objective of this article is to develop an OpenCV-Python code using Haar Cascade algorithm for object and face detection. Currently, UAVs are used for detecting and attacking the infiltrated ground targets. The main drawback for this type of UAVs is that sometimes the object are not properly detected, which thereby causes the object to hit the UAV. This project aims to avoid such unwanted collisions and damages of UAV. UAV is also used for surveillance that uses Voila-jones algorithm to detect and track humans. This algorithm uses cascade object detector function and vision. train function to train the algorithm. The main advantage of this code is the reduced processing time. The Python code was tested with the help of available database of video and image, the output was verified.
\end{abstract}

\begin{IEEEkeywords}
\texttt{Object Detection, Face Detection, Unmanned Aerial Vehicle, Image Processing \& Computer Vision.}
\end{IEEEkeywords}

\section{Introduction}
\lettrine[lines=2]{T}
he Unmanned Aerial Vehicle, which is an aircraft with no pilot on board. UAVs can be remote controlled aircraft (e.g. flown by a pilot at a ground control station) or can fly autonomously based on pre-programmed flight plans or more complex dynamic automation systems. UAVs are currently used for a number of missions, including reconnaissance and attack roles. For the purposes of this article, and to distinguish UAVs from missiles, a UAV is defined as being capable of controlled, sustained level flight and powered by a jet or reciprocating engine. In addition, a cruise missile can be considered to be a UAV, but is treated separately on the basis that the vehicle is the weapon. The acronym UAV has been expanded in some cases to UAVS (Unmanned Aircraft Vehicle System). The FAA has adopted the acronym UAS (Unmanned Aircraft System) to reflect the fact that these complex systems include ground stations and other elements besides the actual air vehicles. Officially, the term 'Unmanned Aerial Vehicle' was changed to 'Unmanned Aircraft System' to reflect the fact that these complex systems include ground stations and other elements besides the actual air vehicles. The term UAS, however, is not widely used as the term UAV has become part of the modern lexicon. As the capabilities grow for all types of UAV, nations continue to subsidize their research and development leading to further advances enabling them to perform a multitude of missions. UAV no longer only perform intelligence, surveillance, and reconnaissance (ISR) missions, although this still remains their predominant type. Their roles have expanded to areas including electronic attack (EA), strike missions, suppression and/or destruction of enemy air defense (SEAD/DEAD), network node or communications relay, combat search and rescue (CSAR), and derivations of these themes.

\section{Computer Vision \& Object Detection }
Today, images and video are everywhere. Online photo sharing sites and social networks have them in the billions. The field of vision research[1,] has been dominated by machine learning and statistics. Using images and video to detect, classify, and track objects or events in order to "understand" a real-world scene. Programming a computer and designing algorithms for understanding what is in these images is the field of computer vision. Computer vision powers applications like image search, robot navigation, medical image analysis, photo management and many more. From a computer vision point of view, the image is a scene consisting of objects of interest and a background represented by everything else in the image. The relations and interactions among these objects are the key factors for scene understanding. Object detection and recognition are two important computer vision tasks. Object detection determines the presence of an object and/or its scope, and locations in the image. Object recognition identifies the object class in the training database, to which the object belongs to. Object detection typically precedes object recognition. It can be treated as a two-class object recognition, where one class represents the object class and another class represents non-object class. Object detection can be further divided into soft detection, which only detects the presence of an object, and hard detection, which detects both the presence and location of the object. Object detection field is typically carried out by searching each part of an image to localize parts, whose photometric or geometric properties match those of the target object in the training database. This can be accomplished by scanning an object template across an image at different locations, scales, and rotations, and a detection is declared if the similarity between the template and the image is sufficiently high. The similarity between a template and an image region can be measured by their correlation (SSD). Over the last several years it has been shown that image based object detectors are sensitive to the training data.

\section{Scope of research}
Image processing based UAV is not completely operational as it is there is a manual intervention of a camera and joy stick. It will reduce the man work time and complexity of the work. In some cases UAVs use very costly laser sensors and multiple sensor integrated systems to detect the objects and people. This project will be useful in replacing the laser sensor and servile the location using cheaper systems. UAV is a very expensive vehicle which cannot be lost under blunders of non-detected objects and unprocessed faces so this project aims in compensating such situations.

\section{Image Processing}
Image processing is a method to convert an image into digital form and perform some operations on it, in order to get an enhanced image or to extract some useful information from it. It is a type of signal dispensation in which input is image, like video frame or photograph and output may be image or characteristics associated with that image. Usually Image Processing system includes treating images as two dimensional signals while applying already set signal processing methods to them. Image processing basically includes the following three steps:

\begin{itemize}
\item Importing the image with optical scanner or by digital photography.
\item Analyzing and manipulating the image which includes data compression and image enhancement and spotting patterns that are not to human eyes like satellite photographs.
\item Output is the last stage in which result can be altered image or report that is basedon image analysis.
\end{itemize}

\subsection{Block Diagram}

\begin{figure}[!h]
\begin{center}
\includegraphics[width=3in]{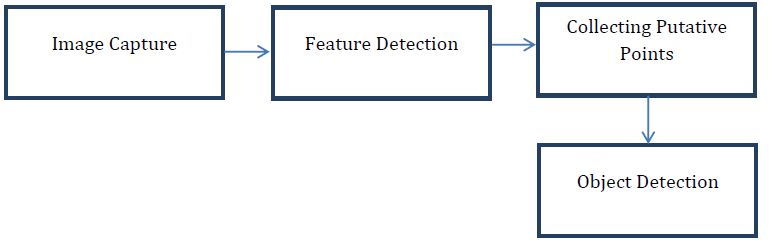}
\caption{Block Diagram for Object Detection}
\label{fig:block}
\end{center}
\end{figure}

Figure 1 shows the image is captured by a camera ${\rightarrow}$ From the image, features are determined by the algorithm ${\rightarrow}$ Form that putative points are collected ${\rightarrow}$ By using the putative points the object to be concreted can be determined from the image.

\section{Results}
\subsection{Face Detection}

A simple face tracking system by dividing the tracking problem into three separate problems:
\begin{itemize}
\item Detect a face to track
\item Identify facial features to track
\item Track the face
\end{itemize}

\vspace{0.7cm}
\large\texttt{Stepwise Procedure:} 
\paragraph{Step 1} \textbf{Detect a Face to Track} 

Before we begin tracking a face, we need to first detect it. Use the vision. Cascade Object Detector to detect the location of a face in a video frame. The cascade object detector uses the Viola-Jones detection algorithm (Later, we will discuss the mathematical modeling of Haar-like features and Viola\&Jones) and a trained classification model for detection . By default, the detector is configured to detect faces, but it can be configured for other object types. The detected face output is shown (see fig. 2).

\vspace*{-0.3cm}

\begin{figure}[!h]
\begin{center}
\includegraphics[width=3in]{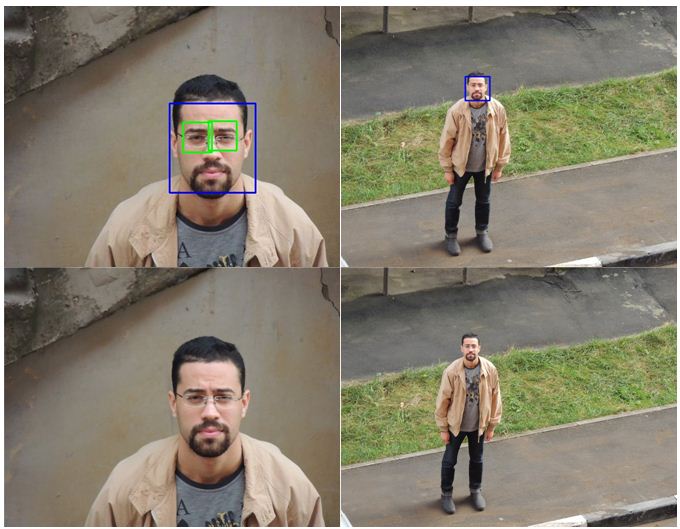}
\caption{Detected Face}
\label{fig:block}
\end{center}
\end{figure}

We can use the cascade object detector to track a face across successive video frames. However, when the face tilts or the person turns their head, we may lose tracking. This limitation is due to the type of trained classification model used for detection. To avoid this issue, and because performing face detection for every video frame is computationally intensive, this example uses a simple facial feature for tracking.\\

\paragraph{Step 2} \textbf{Identify Facial Features to Track}

Once the face is located in the video, the next step is to identify a feature that will help us track the face. For example, we can use the shape, texture, or color. we choose a feature that is unique to the object and remains invariant even when the object moves. In this example, we use color as the feature to track. The color provides a good deal of contrast between the face and the background and does not change as the face rotates or moves (see fig. 3 as an output).

\begin{figure}[!h]
\begin{center}
\includegraphics[width=3in]{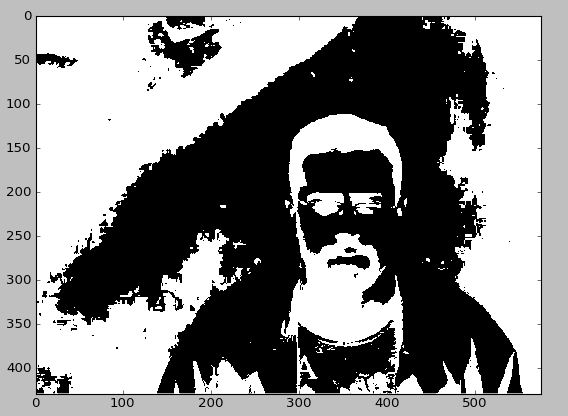}
\caption{Face Tracking Features}
\label{fig:features}
\end{center}
\end{figure}

\vspace*{-0.5cm}
\paragraph{Step 3} \textbf{Track the Face}\\
With the color selected as the feature to track, we can use the vision. Once the face in the video was identified, as a position which occupied in the output area in term of geometric coordinators, we can distinguish between the real face shape and its correspondent background.

\subsection{Object Detection}
In this and previous section, we introduce algorithms to visualize feature spaces used by object detectors. We found that these visualizations allow us to analyze object detection systems in new ways and gain new insight into the detector’s failures. So here we present an algorithm work to detect cars using shape features. The following algorithm presents an algorithm for detecting a specific object based on finding full object (Human body and cars). It can detect objects despite a scale change or a small plan rotation. It is also robust to small amount of out-of-plane rotation and occlusion.

\begin{figure}[!h]
\begin{center}
\includegraphics[width=3.56in]{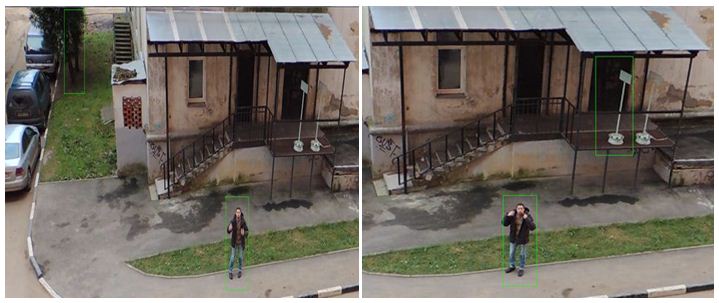}
\caption{Detected Objects}
\label{fig:2}
\end{center}
\end{figure}

\vspace*{-0.5cm}
In figures (fig. 4 and 5) we tried to detect object based on their correspondent shape. It is clear in the figure at the right-hand side (see fig. 4) represents two detected object: a non frontal full human body and a parking sign. The figure at the left-hand side (see fig. 4) represents also two main objects: a frontal full human body and a tree. 

\vspace*{0.5cm}
Let's treat the vehicles case now.
The easy way to do vehicle detection is by using Haar Cascades. To do vehicle tracking, we need to use a tracking algorithm. The Haar Cascades is not the best choice for vehicle tracking because its large number of false positives. What do we mean by False Positives ? In the general case, a false positive is an error in some evaluation process in which a condition tested for is mistakenly found to have been detected.

\begin{figure}[!h]
\begin{center}
\includegraphics[width=3.45in]{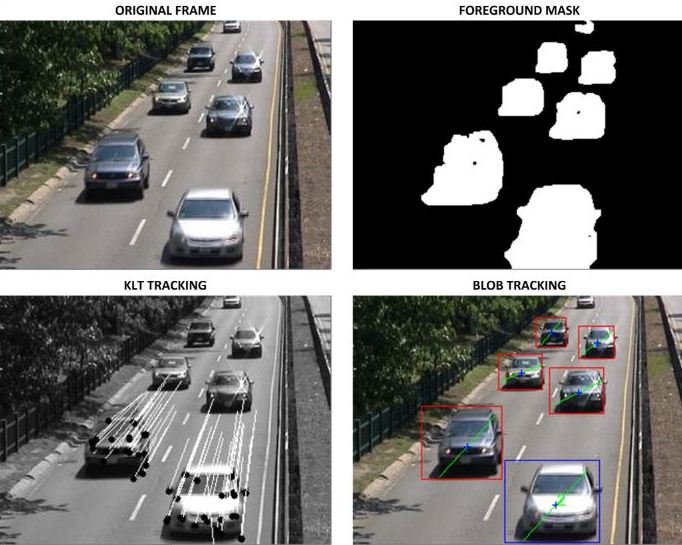}
\caption{Detected Vehicles}
\label{fig:2}
\end{center}
\end{figure}

The Haar-Cascade \textbf{cars3.xml} (see fig. 5) was trained using 526 images of cars from the rear (360x240 pixels, no scale). The images were extracted from the Car dataset proposed by Brad Philip and Paul Updike[6,7,8,9] taken of the freeways of southern California. This algorithm detects any moving object as vehicle. To save the foreground masks, we can use the BGSLibrary[10]. 
One suggestion is to use a BS algorithm. But, only a BS algorithm is insufficient to do vehicle tracking, we will need a blob tracker algorithm or a library like cvBlob or OpenCVBlobsLib. I have taken the origianl video work in (fig. 5 and 6) from youtube.

\begin{figure}[!h]
\begin{center}
\includegraphics[width=3.5in]{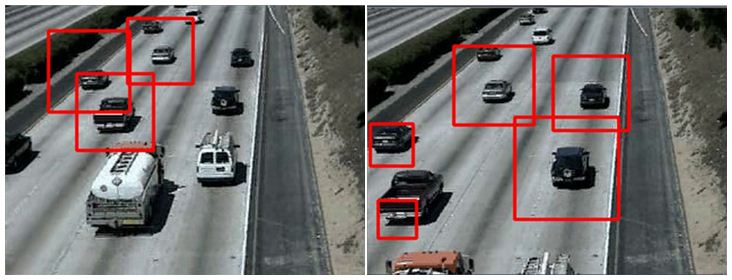}
\caption{Frame Detected Vehicles}
\label{fig:2}
\end{center}
\end{figure}

In order to do some tips to do vehicle tracking and counting, we need to:

\begin{itemize}
\item First, perform a background subtraction.
\item Send the foreground mask to cvBlob or OpenCVBlobsLib.
\item The cvBlob library provide some methods to get the centroid, the track and the ID of the moving objects. We can also set if we want to draw a bounding box, or the centroid and the angle of the tracked object.
\item Check if the centroid of the moving object has crossed a virtual line (or region) in our video.
\end{itemize}

\section{Mathematical Modeling}
\subsection{\textbf{Haar-like features and Viola\&Jones}}
\paragraph{\texttt{Haar Detection Cascade}}
The usage of \textbf{Haar-like features} in object detection[1,2,3,4,5] is proposed first time by Paul Viola and Michael Jones in \textbf{Viola \& Jones (2001)}. This method then attracted so much attention from the others and a lot of people extended this method for a various field of object detection.
The idea behind the Haar Detection Cascade is to eliminate negative examples with very little processing. A series of classifiers are computed to every sub-region in the image. If the sub region does not pass all of the classifiers than that image is released and further computation is performed. If a sub region passes the first stage which requires little computation, it is then passed onto the next stage where a little more computation is needed. If the image passes this classifier it is then passed to another sub region. In order for face detection to occur, an image sub region must pass all of these classifiers. One can train a classifier to improve its accuracy, but the classifier in OpenCV for face detection works just fine for out purpose.

\begin{figure}[!h]
\begin{center}
\includegraphics[width=2.7in]{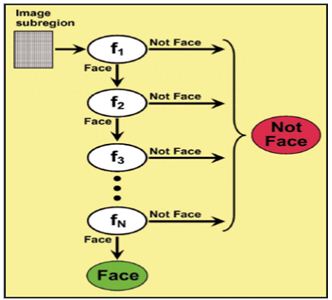}
\caption{Haar Classifier}
\label{fig:Haar Classifier}
\end{center}
\end{figure}

\vspace*{-0.5cm}
Haar-like features are an over complete set of two-dimensional (2D) Haar functions, which can be used to encode local appearance of objects[11]. They consist of two or more rectangular regions enclosed in a template. The feature value f of a Haar-like feature which has k rectangles is obtained as in the following equation:

\begin{equation}
f = \sum_{i=1}^{k} w^{(i)}. \mu^{(i)} 
\end{equation}

where $\mu^{(i)}$ is the mean intensity of the pixels in image \textbf{x} enclosed by the ith rectangle. The quantity $\mu$ refers as the rectangle mean. In (1), $\mu^{(i)}$ is the weight assigned to the $i^{th}$ rectangle. Traditionally, the weights assigned to the rectangles of a Haar-like feature are set to default integer numbers such that the following equation is satisfied: 

\begin{equation}
\sum_{i=1}^{k} w^{(i)} = 0
\end{equation}

One of the main reasons for the popularity of the Haar-like features is that they provide a very attractive trade-off between speed of evaluation and accuracy. With a simple weak classifier based on Haar-like features costing just 60 microprocessor instructions, Viola and Jones[14] achieved 1\%	 false negatives and 40\% false positives for the face detection problem. The high speed of evaluation is mainly due to the use of integral images[14], which once computed, can be used to rapidly evaluate any Haar-like feature at any scale in constant time. Since the introduction of horizontally and vertically aligned Haar-like features by Papageogiou et al.[11], many different Haar-like features have appeared in the literature[12,13,14]. The main difference in the Haar-like features concept is in the number of rectangles and the orientation of the rectangles with respect to the template. Jones and Viola[15] introduced diagonal features to capture diagonal structures in object's appearance. Lienhart[13] enriched the features of Viola and Jones [14] with efficiently computable rotated Haar-like features. With the enriched set, they achieve a 10\% lower false alarm rate for a given true positive rate. Li et al.[12] proposed Haar-like features with disjoint rectangles which were meant to characterize
non-symmetrical features of an object's appearance.

\vspace*{0.5cm}
\paragraph{\texttt{The Viola Jones framework}}
The work of Viola and Jones [3,16] can be considered one of the first robust real time face detectors that is still being used in practice. \underline{\textit{Three}} main ideas are behind the success of this detector: \textbf{the integral image}, \textbf{the adaboost} and \textbf{the attentional cascade structure}[17].

\vspace*{0.5cm} 
\underline{\textbf{The integral image}}[3] is an algorithm for a quick and efficient calculation for the sum of intensity values in a rectangular subset of an image. The integral image is defined as:

\begin{equation}
ii(x,y)= \sum_{\substack{
x'\le x \\
y'\le y
}} i(x',y')
\end{equation}

where i(x, y) is the intensity of the gray scale image at pixel (x, y). Using the integral image as illustrated in Fig. 8, the sum of the intensity pixels of any rectangular area ABCD can be calculated with only four array references as: 

\begin{small}
\begin{equation}
\sum_{(x,y)\in ABCD}^{} i(x,y) = ii(D) + ii(A) - ii (B) - ii(C) 
\end{equation}
\end{small}

Viola and Jones used this concept for rapid computation of a huge number of Haar like features which are simply defined as the difference between the sum of the intensities in
the dark and light shaded regions of simple rectangular patterns as shown in Fig. 8(e). 

\begin{figure}[!h]
\includegraphics[width=3.55in]{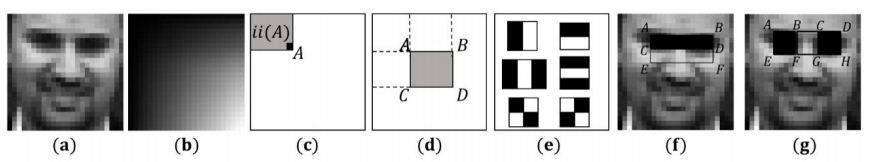}
\captionsetup{singlelinecheck=off,font=footnotesize}
\caption{Efficient calculation of Haar like features using integral image: (a) Original image (b) Integral image (c) illustration of ii(A) (d) Illustration of equation 4 (e) Examples of Haar like features (f) Example of a Haar feature overlapped with original image and illustration of equation 5 (g) Example of another Haar feature overlapped with original image.}
\end{figure}

For example, the feature shown in Fig 8(f) can be calculated as:

\begin{scriptsize}
\begin{equation}
f = [ii(D) + ii(A) - ii (B) - ii(C)] - [ii(F) + ii(C) - ii (D) - ii(E)]
\end{equation}
\end{scriptsize}

This computational advantage enabled scaling the features for multi-scale detection at no additional cost because it requires the same number of operations despite of size. In contrast to the conventional image pyramid used in most face detectors to detect over multiscales by scanning a fixed scale detector over different scales of the image, Viola and Jones scaled the detector itself and saved the time of building the image pyramid.

\vspace*{0.5cm}
\underline{\textbf{The Adaboost}} is used both to select features and to train the classifier. The weak learner is designed here to select the feature which best separates the weighted positive and negative training examples. A weak classifier h(x,f,p,$\theta$) is defined as: 

\begin{equation}
h(x,f,p,\theta) = \begin{cases} 1 & \mbox{if } pf(x)<\mbox{p$\theta$} \\ 0 & \mbox{otherwise}  \end{cases}
\end{equation}

Where f is a feature from the huge set spanning different sizes of the Haar like features shown in Figure 8(e), p is a polarity indicating the direction of the inequality, $\theta$ is a threshold and x is a training sub window of size 24x24 pixel. This weak classifiers that threshold single features can be viewed as single node decision trees which are usually called decision stups in the machine learning literature. The Adaboost algorithm of Viola and Jones is described in[17].

\vspace*{0.5cm}
\underline{\textbf{The attentional cascade}}[3] of classifiers is used to combine increasingly more complex classifiers successively which allows background regions of the image to be quickly discarded while spending more computation on promising object-like regions. Simpler and therefore faster boosted classifiers (with low T) are used first to reject the majority of negative windows while passing almost all positive windows. Then more complex and therefore slower boosted classifiers (with high T) are used to reject the much fewer number of difficult negative windows.

\section{Discussion \& Conclusion}
We have develped this article from general to private, means from the need of computer vision to how and why to detect objects and faces. We explained in detailed all concepts of object and face detection and why is it so important that field ? Our results showed that the main aim was to detect the objects and the output objects were detected from the real scene. The face detection program can be implemented to detect and follow people in case of surveillance and other domains. We introduced a tool to explain some of the success of object detection systems. We present algorithms to visualize the success spaces of object detectors. Our results are supported by those of Paul Viola and Michael Jones (2001)[14]. We presented and evaluated also, several algorithms to visualize object detection features to confirm results in[14,2].\\

This work is done in Python-OpenCV and can be performed with Matlab also but we prefer Python because we can include it in OpenCV programs and the execution time in Python is lesser and simple. I conclude my project report by pointing it use in surveillance and obstacle detection process. In future this program can be used to control the cameras in a UAV and navigate through obstacles effectively. This program can be upgraded to reduce the process time of the controller so a different methodology can be tried and implemented. We also believe that visualizations can be a powerful tool for understanding object detection systems and advancing research in computer vision. To this end, We hope more intuitive visualizations will prove useful for the community.

\end{document}